\documentclass[10pt,journal,compsoc]{IEEEtran}
\newif\ifpeerreview

\peerreviewfalse

\usepackage[nocompress]{cite}
\usepackage{url}
\usepackage{amsmath,amssymb,graphicx}

\usepackage{lipsum} 

\usepackage[switch]{lineno}

\usepackage{amsmath}
\usepackage{amssymb}
\usepackage{booktabs}
\usepackage{caption,setspace}
\usepackage{multirow}
\usepackage{float}
\newcommand{\tabincell}[2]{\begin{tabular}{@{}#1@{}}#2\end{tabular}}  
\usepackage[pagebackref,breaklinks,colorlinks]{hyperref}
\usepackage{arydshln}
\usepackage{xcolor}

\usepackage[capitalize]{cleveref}
\crefname{section}{Sec.}{Secs.}
\Crefname{section}{Section}{Sections}
\Crefname{table}{Table}{Tables}
\crefname{table}{Tab.}{Tabs.}

\newcommand{\bI}{{\bf I}}
\newcommand{\bF}{{\bf F}}
\newcommand{\bT}{{\bf T}}
\newcommand{\bC}{{\bf C}}
\newcommand{\bff}{{\bf f}}

\newcommand{\bomega}{{\boldsymbol \omega}}
\newcommand{\bbeta}{{\boldsymbol \beta}}
\newcommand{\bepsilon}{{\boldsymbol \epsilon}}
\newcommand{\bp}{{\bf p}}
\newcommand{\bv}{{\bf v}}
\newcommand{\br}{{\bf r}}
\newcommand{\bd}{{\bf d}}
\newcommand{\bo}{{\bf o}}
\newcommand{\bc}{{\bf c}}
\newcommand{\bx}{{\bf x}}
\newcommand{\by}{{\bf y}}
\newcommand{\bh}{{\bf h}}
\newcommand{\bn}{{\bf n}}
\newcommand{\ie}{\textit{i}.\textit{e}.}
\newcommand{\eg}{\textit{e}.\textit{g}.}

\usepackage[misc]{ifsym}

\usepackage[hang]{footmisc}

\makeatletter
\def\@IEEEsectpunct{.\ \,}

\makeatother

\graphicspath{{images/}}

\newcommand{\annotated}{}

\title{MPS-NeRF: Generalizable 3D Human Rendering from Multiview Images}

\author{Xiangjun~Gao, Jiaolong~Yang, Jongyoo~Kim, Sida~Peng, Zicheng~Liu, and Xin~Tong
\IEEEcompsocitemizethanks{
\IEEEcompsocthanksitem X. Gao is with Beijing Laboratory of Intelligent Information Technology, School of Computer Science, Beijing Institute of Technology, Beijing, 100081, China. Work done during internship at Microsoft Research Asia. Email:  xiangjun\_gao@bit.edu.cn
\IEEEcompsocthanksitem J. Yang, J. Kim and X. Tong are with Microsoft Research Asia, Beijing, 100080, China. J. Yang is the corresponding author of this paper. Email: \{jiaoyan, jongk, xtong\}@microsoft.com
\IEEEcompsocthanksitem S. Peng is with College of Computer Science and Technology, Zhejiang University, Hangzhou, Zhejiang, China. Email: pengsida@zju.edu.cn
\IEEEcompsocthanksitem Z. Liu is with Microsoft Azure AI, Redmond, WA, USA. Email: zliu@microsoft.com
}
}

\begin{document}

\IEEEtitleabstractindextext{%
\begin{abstract}
There has been rapid progress recently on 3D human rendering, including novel view synthesis and pose animation, 
based on the advances of neural radiance fields (NeRF). However, most existing methods focus on person-specific training and their training typically requires multi-view videos. This paper deals with a new challenging task -- rendering novel views and novel poses for a person \emph{unseen} in training, using only multiview \emph{still images} as input without videos. For this task, we propose a simple yet surprisingly effective method to train a generalizable NeRF with multiview images as conditional input. The key ingredient is a dedicated representation combining a canonical NeRF and a volume deformation scheme. Using a canonical space enables our method to learn shared properties of human and easily generalize to different people. Volume deformation is used to connect the canonical space with input and target images and query image features for radiance and density prediction. 
We leverage the parametric 3D human model fitted on the input images to derive the deformation, which works quite well in practice when combined with our canonical NeRF. The experiments on both real and synthetic data with the novel view synthesis and pose animation tasks collectively demonstrate the efficacy of our method.
\end{abstract}

}

\maketitle
\thispagestyle{empty}

\IEEEraisesectionheading{
  \section{Introduction}\label{sec:introduction}
}

\begin{figure*}[!t]
	\centering
	\includegraphics[width=1\linewidth]{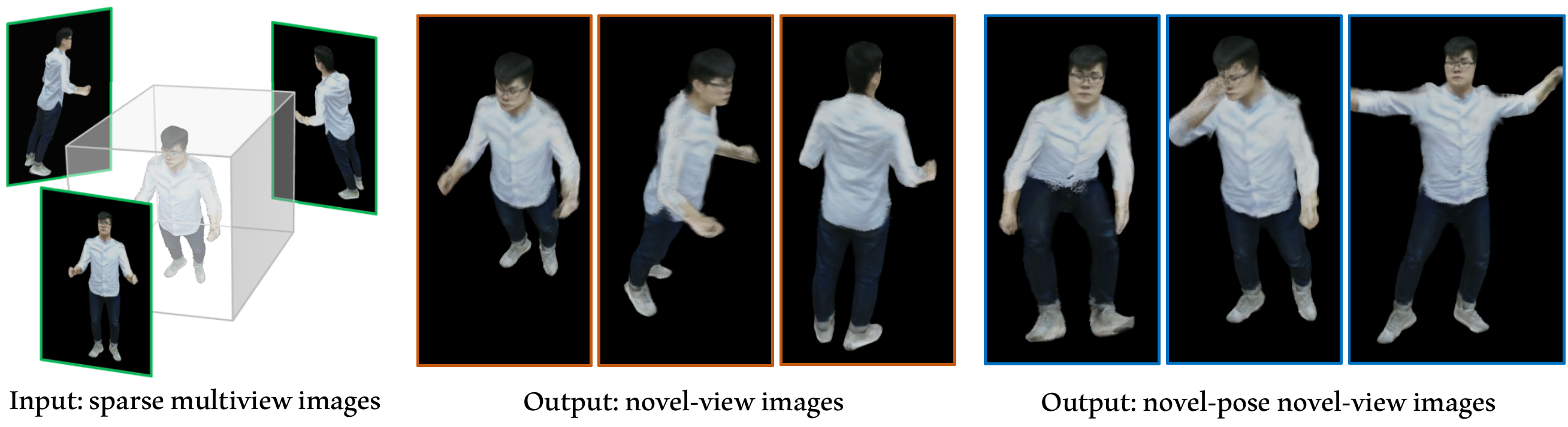}
	\caption{Given a sparse set of multiview \textbf{\emph{images}} of an \textbf{\emph{unseen}} person, our method is able to synthesize novel views of the person and animate it with novel poses. It works in a feed-forward fashion without any optimization.\label{fig:teaser}}
\end{figure*}

\IEEEPARstart{F}{ree-view} human character rendering and animation has numerous applications in avatar creation, telepresence, movie production, among others. While traditional methods~\cite{collet2015high,orts2016holoportation,martin2018lookingood} use dense multiview camera rigs or depth sensors to accomplish this task, recent neural rendering approaches~\cite{wu2020multi,peng2021neural,peng2021animatable,noguchi2021neural} have shown that free-view rendering and animation can be achieved using sparse color cameras, which could significantly reduce the device setup and capture cost. In particular, promising results have been shown by methods~\cite{peng2021neural,peng2021animatable,noguchi2021neural} that are based on the neural radiance field (NeRF)~\cite{mildenhall2020nerf} representation. With NeRF, the generated multiview images not only have decent quality, but also enjoy high 3D consistency by virtue of an explicit, physics-based rendering process, thus enabling visually pleasing free-view and animated human video creation.

However, due to the high complexity of human motion and appearance, existing methods~\cite{peng2021neural,peng2021animatable,noguchi2021neural} are typically trained in a person-specific setup, \ie, one model is trained for one single person, to obtain best rendering quality. Such a setup is clearly not scalable, as rendering any new character would require a  tedious model training process. Moreover, these methods needs multiview \emph{video} for training in order to handle different human poses. This requirement further restricts the practical usefulness of these methods. 

This paper deals with a challenging but more practical problem setup -- \emph{training a model that can render unseen persons directly in a feed-forward manner, and using only still images captured at sparse viewpoints as input}, as illustrated in Fig.~\ref{fig:teaser}. A simple yet effective method is proposed for this task. One key insight is that human bodies share a similar geometric structure, which can be leveraged to learn a rendering model transferable to new subjects. This prior knowledge has been explored in different domains for multi-person 3D representation \cite{anguelov2005scape, ma2019learning, tiwari2021neural, chen2022gdna} and should apply to our radiance field learning as well.  Besides, prior studies~\cite{yu2021pixelnerf,trevithick2021grf} have shown that using images as conditional input, it is possible to train a generic NeRF for common objects, especially for those in a same category. Although these methods focus on rigid objects or static scenes, the spirit of generalization applies to our problem as well.

To leverage the shared human body structure to attain better generalization capability, we learn a radiance field in a canonical space where different people with various poses are well aligned. We use a pose-aligned canonical space defined by a 3D human body parametric model -- SMPL~\cite{loper2015smpl} in this work -- to achieve this. We apply {pre-defined} volumetric deformations to connect the canonical space with 
the target space (for the target image) and the observation space (for the input images).
The deformation is defined by propagating the skinning weights defined on the SMPL human surface to the 3D volume and applying standard linear blend skinning. 
The input to NeRF is a 3D point in the canonical space and corresponding image features retrieved on the input images, similar to \cite{yu2021pixelnerf,trevithick2021grf}. 

Although the deformation so-obtained is simple and cannot perfectly model all complex motions, our rendering scheme is volumetric rendering (as opposed to direct mesh rendering) and is learned to achieve best image rendering quality to the extent possible. The image feature extraction and radiance prediction networks will adapt to such a deformation scheme through the training process.

Our method is named Multi-Person Skinning NeRF, or MPS-NeRF, for the goal to handle generic person and its skinning-based deformation derived by SMPL~\cite{loper2015smpl}. It allows for not only novel view synthesis but also plausible novel pose animation, thanks to the animatable nature of our deformation-based NeRF representation and its seamless combination with the parametric human model. We evaluate our method on the Human3.6M dataset~\cite{ionescu2013human3} and another synthetic dataset 
created by the body models from the THuman~\cite{zheng2019deephuman} dataset.
The results on both novel-view and novel-pose synthesis tasks have demonstrated the effectiveness of our method.

\textbf{The contributions of this paper} can be summarized as follows. First, we explore a challenging task of novel-view and novel-pose rendering for unseen persons given only sparse multiview images. To our knowledge, this task is not handled by any existing method. We propose a novel and simple method for this task, which defines a shared canonical space to facilitate generalizable NeRF training and leverages a parametric human model to derive volume deformation. We apply carefully-designed losses to fully leverage the priors of the parametric human model to achieve generalization to unseen person. Albeit the conceptual simplicity of our method, it works surprisingly well as demonstrated on both real and synthetic datasets. Somewhat counter-intuitively, the rendering quality is even on par with or better than previous person-specific models. Our method could serve as a strong baseline model for future works towards generalizable 3D human rendering.

\section{Related Work}
\textbf{Neural 3D representations and rendering.}
Recently, neural rendering based on implicit function has emerged as an effective way of novel view synthesis.
In this context, various 3D representations were studied such as occupancy networks~\cite{mescheder2019occupancy}, SDF~\cite{park2019deepsdf}, Implicit field~\cite{chen2019learning}, and SRN~\cite{sitzmann2019srns}.
In particular, NeRF~\cite{mildenhall2020nerf} achieves photorealistic synthesis by modeling the 3D scene as a continuous 5D function.
Though NeRF yields impressive quality, it can only deal with static scenes. Several studies try to handle non-rigid objects by employing deformation to map the observation space to the canonical space \cite{pumarola2021d, park2021nerfies, tretschk2021nonrigid}. Some works are devoted to generalizing NeRF with images as conditional input~\cite{yu2021pixelnerf,trevithick2021grf,wang2021ibrnet}. This paper presents a NeRF-based method dedicated to generalizable human rendering.

\vspace{4pt}
\noindent\textbf{Human performance capture and 3D reconstruction.}
3D human reconstruction has been widely studied in the literature. 
Earlier works are based on 2D keypoints \cite{gall2009motion}, multi-view consistency \cite{de2008performance}, or depth maps \cite{bogo2015detailed, wei2012accurate}.
Later, parametric 3D human models \cite{loper2015smpl, joo2018total} have been frequently deployed
\cite{bogo2016keep, Lassner2017up, kolotouros2019cmr, kolotouros2019spin, Pavlakos2019expressive}, where optimal parameters which deforms the human model to match the input data are fitted. However, the parametric 3D model has the limited capacity, leading to less generalizability to clothed people.
Recently, implicit field based methods were proposed to capture detailed surface geometry from one or few views \cite{saito2019pifu, saito2020pifuhd}. Yet, the reconstructed geometry is difficult to be deformed into novel poses.
ARCH~\cite{huang2020arch} and ARCH++~\cite{he2021arch++} combine the parametric model with the implicit field to predict rigged 3D clothed humans. The model yields the detailed geometry as well as the pixel-aligned colors.
However, these approaches
require large-scale ground-truth meshes of clothed persons, while ours is trained on multiview images without any 3D supervision.

\vspace{4pt}
\noindent\textbf{Human rendering.}
A few studies have been proposed for human image synthesis with the NeRF framework.
Generally, human-prior information such as a skeleton or a parametric model was adopted to mitigate the large deformation of human bodies.
NeuralBody~\cite{peng2021neural} adopts SMPL~\cite{loper2015smpl} and uses per-vertex latent code which is used to generate a continuous latent code volume. 
By deforming the SMPL mesh, any pose and view can be rendered through the proposed framework.

\textcolor{black}{Some methods \cite{liu2021neural, peng2021animatable, weng2022humannerf} also leverage SMPL to deform the 3D space to a canonical pose, where pose-dependent residual deformation can be considered.  H-NeRF~\cite{xu2021h} combines SDF with NeRF for temporal rendering and reconstruction. 
{\annotated A-NeRF~\cite{su2021nerf} uses 3D articulated skeleton pose to build an animatable NeRF model from a monocular video.}
Although these method can synthesize plausible images, they are designed for the person-specific setup.
\cite{kwon2021neural} proposes a generalizable method that handles unseen person, but they still need multiview videos and do not address still image input. A recent work of  \cite{zhao2021humannerf} considers generalizable human rendering using images as input, but it does not handle novel pose rendering. 
In this work, we aim at learning a generalizable model to achieve both novel view and novel pose rendering, using only multiview images as input.}

\begin{figure*}[th!]
	\centering
	\includegraphics[width=1.0\linewidth]{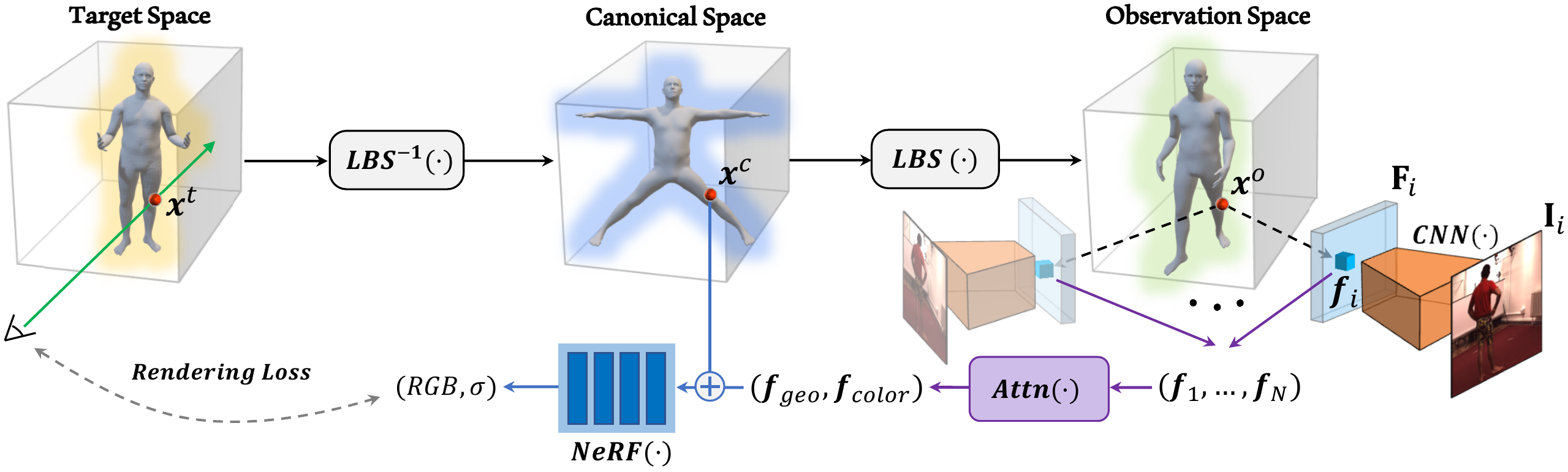}
	\vspace{-10pt}
	\captionsetup{type=figure,font=small}
	\caption{
		\textbf{Method Overview.} \textcolor{black}{To render the target image, we first cast rays and sample points in target space. Then each sampled point $\bx^t$ is deformed to the canonical space by inverse linear blend skinning and the deformed point is used as the input to NeRF. The point is further deformed to the observation space and projected onto the input images to retrieve multiview features which are fused by self-attention blocks. The fused features are then used for density and radiance prediction. \label{fig:framework}}
	}
	\vspace{0pt}
\end{figure*}

\section{Proposed Method}
Ours goal is to train a generic model which can directly synthesize a human image with novel viewpoint and/or pose in a feed-forward fashion using only multiview images as conditional input.
For the input multiview images, we assume the calibration parameters and the human region masks are known. 
We also assume the parameters of a 3D human parametric models fitted to the multiview images are given. In this work, we use SMPL~\cite{loper2015smpl} as our parametric model. The SMPL parameters can be obtained by applying existing fitting methods based on keypoints, silhouette, multiview consistency or any other possible means~\cite{bogo2016keep,joo2018total,zhang2020object,li20213d}.

\subsection{Overview}
The overall framework of our method is presented in Fig.~\ref{fig:framework}. The input consists of a sparse set of observed multiview images $\{\bI_i\}$ associated with their camera parameters $\{\bv^o_i\}$ and fitted SMPL model shape and pose parameters $\bbeta, \bp^o$, and the camera and pose parameters $\bv^t,\bp^t$ of the target view. The output is the human image rendered at $\bv^t$ with pose $\bp^t$. Note that if $\bp^t=\bp^o$, the task is known as novel view synthesis; otherwise we synthesize novel poses as in an animation task.

To render the target image, we follow recent works~\cite{peng2021neural,peng2021animatable,noguchi2021neural} and base our rendering scheme on NeRF~\cite{mildenhall2020nerf}, which is a compact yet powerful representation for neural rendering. To compute the color of a pixel on the target image, we cast a ray to the 3D space which passes the camera center and the pixel. We then sample points along the ray, predict their densities and colors using a neural network, and accumulate the colors following the volumetric rendering scheme~\cite{max1995optical,mildenhall2020nerf}.

The key lies in how we compute the density $\sigma$ and color $\bc$ for a point $\bx^t$ in the target space. Inspired by \cite{yu2021pixelnerf,trevithick2021grf}, we use image features as conditional input to the network for density and color prediction.
A straightforward solution would be projecting $\bx^t$ onto the input image planes to get features and then combining the features with $\bx^t$ as the input to NeRF. However, the complex, non-rigid human motions make the radiance field prediction difficult to learn and generalize to different people with various poses.
Besides, this method can only be applied to the novel view synthesis task. For novel pose synthesis, it lacks a way to handle pose change for feature retrieval, nor does it offer a mechanism to drive an animation with desired pose change.  

Our method defines a canonical space where 3D human bodies are aligned based on the SMPL model. To predict the density and color, we deform $\bx^t$ to this canonical space and use the deformed point as the input to the network instead of $\bx^t$. The point is further deformed to the observation space and projected onto the input images to retrieve the multiview features which are also used for radiance prediction.

\subsection{Canonical Space and Volume Deformation}

The canonical space is a pose-aligned space for different human bodies. Its coordinate system is defined the same as the SMPL model. 
The pose shown in Fig.~\ref{fig:framework} is defined as our canonical pose, and we found that using this canonical pose leads to better geometry and rendering results for the leg region compared to the T-shape rest pose in SMPL. 
The notion of a canonical space has been used in previous deformable NeRF schemes such as \cite{pumarola2021d,park2021nerfies,peng2021animatable}. Compared to these methods, our main motivation is to learn a NeRF for different subjects in one shared space to foster generalization, whereas their goal is to model the dynamics of one single scene or object.

We apply two deformation fields to connect the canonical space with the observation space and target space, respectively. The former is used to retrieve image features from the input images for radiance prediction, while the latter is for rendering the output image. 
We formulate our deformation as an extended skinning process by propagating the surface skinning weights of SMPL model to the volume. The SMPL model provides a pre-defined skinning weight vector $\bomega \in \mathbb{R}^{24}$ for each vertex on the body surface for skinning. Following \cite{huang2020arch,bhatnagar2020loopreg,peng2021animatable}, for each point in the volume, we assign the skinning weights of its closest body vertex. 
Note that nothing prevents us from using multiple nearest neighbors on body surface to build a fuzzy association similar to \cite{yang2018analyzing}. In practice, we did not observe significant improvements in our experiments and thus opted for the simplest formulation.
With the assigned skinning weights, volume deformation can be calculated by the linear blend skinning (LBS) algorithm~\cite{lewis2000pose}. Specifically, for a point $\bx^c$ in the canonical space, the function $\rho^{c\rightarrow o}(\cdot)$ deforming it to the observation space can be written as
\begin{equation}
	\bx^o\!=\!\rho^{c\rightarrow o}(\bx^c)\!=\!LBS\big(\bx^c\!, \bomega(\bx^c)\big)\!=\!\Big(\!\sum_{j=1}^{24} \omega_j(\bx^c) \bT_j\!\Big) \bx^c\!,
\end{equation}
where $\bT_j\in \mathrm{SE}(3), j=1,\ldots,24$ are the known rigid transformations of the body joints.
Deforming a point $\bx^t$ in the target space to the canonical space requires an inverse LBS function:
\begin{equation}
	\bx^c\!=\!\rho^{t\rightarrow c}(\bx^t)\! = \!LBS^{-\!1}\big(\bx^t\!, \bomega(\bx^t)\big)\!=\!\Big(\!\sum_{j=1}^{24} \omega_j(\bx^t) \bT_j\!\Big)^{\!\!-\!1}\!\bx^t\!.
	\raisetag{7pt}
\end{equation}

\vspace{4pt}
\noindent\textbf{Discussion.}
The deformations so-obtained provide an approximation of the true correspondence field when the target and observed poses are different (\ie, $\bp^t\ne\bp^o$).
{\annotated They should be reasonably accurate for regions nearby the SMPL surface. For distant regions, the density will be predicted as zero so the accuracy of deformation does not influence rendering.}
Prior works shows that it is possible to learn a refined deformation field in some constrained situation (\eg, a residual skinning weight field is learned in the person-specific model of \cite{peng2021animatable}). However, this task is extremely challenging for the inverse LBS, as a point $\bx^t$ in the 3D space could be on arbitrary body part for different target poses. In \cite{peng2021animatable}, a per-frame latent code is jointly learned to alleviate this issue, and a test-time optimization is further needed to optimize the latent code for a novel pose. However, learning such a deformation network that is generalizable to any unseen person under arbitrary pose is prohibitively difficult.

Our finding is that the simple deformation scheme works quite well in our method, where an image-conditioned NeRF is trained with this deformation scheme and will learn to adapt to it to ensure best rendering quality.

\subsection{Image-Conditioned Rendering}

The input to our canonical NeRF is a point in  the canonical space and features extracted from the input images. To get the image features, we first apply a CNN on the images $\{\bI_i\}$ to extract feature maps $\{\bF_i\}$. For a canonical space point $\bx^c$, we deform it to the observation space as $\bx^o$ and project it onto the input image planes. A set of features is extracted using bilinear interpolation. We also sample RGB colors on the images and append them to the extracted features to form the final image features $\{\bff_i\}$:
\begin{equation}
	\bff_i  = \big[\bF_i\big(\Pi(\bx^o,\bv^o_i)), \bI_i\big(\Pi(\bx^o,\bv^o_i)\big)\big],
\end{equation}
where $\Pi(\cdot)$ denotes the 3D-2D projection function.

Next, we fuse the extracted multiview features for the subsequent radiance field prediction. Human images contain severe self-occlusions, thus feature fusion is not trivial especially under very few views (\eg, three or four) with large view angle differences. In our method, we employ the attention mechanism in Transformers~\cite{vaswani2017attention,dosovitskiy2020image} for effective feature fusion. Two self-attention blocks are applied to generate two features, one for geometry prediction and another for color:
\begin{equation}
	\bff_{geo}\!=\!Attn_{geo}\big(\{\bff_i\}\big),~ \bff_{color}\!=\!Attn_{color}\big(\{\bff_i\}\big).
\end{equation}

The fused features, together with the coordinate of canonical space point, are then fed into an MLP network to predict density $\sigma$ and color $\bc$. Our network is adapted from \cite{mildenhall2020nerf} with a few tweaks. Notably, we replace the view direction input with our fused color feature for the final color prediction:
\begin{equation}
\begin{split}
	\sigma, \bh &= M~\!\!L~\!\!P_1(\gamma(\bx^c), \bff_{geo}), \\
	\bc &= M~\!\!L~\!\!P_2(\bh, \bff_{color}),
\end{split}
\end{equation}
where $M~\!\!L~\!\!P_1$ and $M~\!\!L~\!\!P_2$ are partial MLP layers with same structures as \cite{mildenhall2020nerf} except for the first layer, and $\gamma(\cdot)$ is the positional encoding function.

{\annotated
Note that attention-based multiview feature fusion has been used in some previous methods such as \cite{trevithick2021grf,wang2021ibrnet,kwon2021neural}. Our method is different from them in that we designed two attention MLPs to generate two features for geometry and color, respectively. Generating two features should not only increase the representation power but also ease the feature fusion learning process. Empirically, we also find this design leads to better performance than using one feature fused with a single attention block, as will be shown in the experiments.
}

Finally, the volumetric rendering procedure~\cite{kajiya1984ray,mildenhall2020nerf} for a ray $\br(t)=\bo+t\bd$ in the target space can be written as: 
\begin{equation}
\begin{split}
	\bC(\br)&=\int^{t_f}_{t_n} T(t)\sigma(\rho^{t\rightarrow c}(\br(t)))\bc(\rho^{t\rightarrow c}(\br(t)))dt,\\
	T(t)&={\exp}\Big(-\int^t_{t_n} \sigma(\rho^{t\rightarrow c}(\br(s)))ds\Big),
\end{split}
\end{equation}
where the integrals can be estimated by using point samples along the ray.

\subsection{Training Loss}

Our method contains three submodules that need to be trained: the CNN for image feature encoding, the attention blocks for feature fusion, and the MLP for canonical NeRF. All these networks are trained in an end-to-end manner using images from a collection of people in the training set. After training, our method can be applied on new subjects for novel view and novel pose synthesis. The following loss functions are used for training.

\vspace{4pt}
\noindent\textbf{Color loss.}
Given ground-truth target images, we apply the color loss to supervise the training, defined as:
\begin{equation}
	L_{color} =  \sum_{\br\in\Omega} \| \bC (\br)-\bC^*(\br) \|^2_{2},
\end{equation}
where $\Omega$ is the ray collections for pixels on the target image, and $\bC^*(\br)$ is the ground-truth color.

\vspace{4pt}
\noindent\textbf{Mask loss.}
We also leverage the mask labels of the ground-truth images to train the radiance fields. The mask loss is defined as:
\begin{equation}
	L_{mask}= \sum_{\br\in\Omega} \| M(\br) -  M^*(\br) \|^2_{2},
\end{equation}
where $M(\br)$ is the accumulated volume density and $M^*(\br)$ is the binary mask label.

{\annotated To better leverage the geometry prior from the SMPL model, we further impose the following two new losses.}

\vspace{4pt}
\noindent\textbf{Smoothness loss.}
We incorporate a normal smoothness prior to encourage smooth geometry. Similar to \cite{oechsle2021unisurf}, we enforce the normal vector of a canonical point $\bx^c$ and a point sampled in its neighborhood to be close:
\begin{equation}
	L_{smooth} =  \sum_{\bx^c} \| \bn(\bx^c) -\bn(\bx^c+\bepsilon) \|^2_{2},
\end{equation}
where $\bepsilon$ is a small perturbation randomly drawn from uniform distribution during training. The normal at a point $\bx^c$ is calculated  by $\bn(\bx^c) = \frac{\nabla_{\bx^c} \sigma(\bx^c)}{\Vert \nabla_{\bx^c} \sigma(\bx^c) \Vert _{2}}$ where $\nabla$ denotes the spacial gradient which can be obtained by network backpropagation. 

\vspace{4pt}
\noindent\textbf{Shape loss.} To further regularize the learned geometry and avoid overfitting, we add a weak constraint that the learned geometry should not be too far away from the fitting SMPL model. Concretely, we penalize the normal of a point $\bx^c$ in our learned geometry and that of its nearest body vertex $\by$ on the SMPL body surface:
\begin{equation}
	\label{e11}
	L_{shape} =  \sum_{\bx_c \in \Omega} \Vert \bn(\bx^c) -\bn(\by^c) \Vert^2_{2},
\end{equation}
where SMPL vertex normal $\bn(\by^c)$ is constant and can be pre-computed. Here we sample $\bx^c$ by adding small random perturbations to the ray sampling points that are within a distance threshold to SMPL surface.

In summary, the overall loss function we use to train our MPS-NeRF can be written as:
\begin{equation}
	L=L_{color} + \lambda_{1} L_{mask} + \lambda_{2} L_{smooth} + \lambda_{3} L_{shape},
\end{equation}
where $\lambda$'s are the balancing weights.

\section{Experimental Results}

\noindent\textbf{Implementation details.}
Our method is implemented with PyTorch\footnote{All source codes and trained models will be publicly released.}. 
We use Adam optimizer~\cite{kingma2015adam} with a learning rate of ${5e-4}$ to train the models.
The loss weights are set as $\lambda_{1}{=}1.0$, $\lambda_{2}{=}0.1$, and $\lambda_{3}{=}0.1$ in all the experiments.
We use 2 Nvidia Tesla V100 GPUs for training with a batch size of 350 rays on each GPU. Training takes about two days on the datasets we used.

We use the first 7 convolution layers of ResNet-34~\cite{he2016deep} as our image feature encoder, with the original max-pooling layer removed. The images are downsampled to half resolution before feeding into the encoder CNN. The MLP for canonical NeRF is adapted from \cite{mildenhall2020nerf}. The $MLP_1$ subnet (Eq.~5) has 8 layers each with 256 feature dimensions.
$MLP_2$ (Eq.~6) has 2 hidden layers with 256 feature dimensions. ReLU activation is used for all hidden layers. Following Mip-NeRF~\cite{barron2021mip},  we use a shifted softplus activation: $\log(1+ \exp(x-1))$ to produce density $\sigma$.  For color $\bc$,  instead of using the sigmoid activation, we use a ``widened" sigmoid function $f(x) = (1+2\epsilon) / (1+\exp(-x)) - \epsilon$, with $\epsilon=0.001$.

To render a pixel on the target image, a ray is cast to the 3D volume and points are sampled along the ray for radiance accumulation.
Instead of using the sampling strategy in original NeRF~\cite{mildenhall2020nerf} which samples points in the whole volume, we follow \cite{peng2021neural} to sample points within a 3D bounding box derived based on the SMPL model. 

Our learning rate starts from $5 e {-4}$ and decays by a factor $0.5$ for every $30$K interactions, and the maximum iteration number is set to $120$K.

\vspace{4pt}
\noindent\textbf{Datasets.}
We mainly use two datasets to evaluated our method and compare it with previous methods. The first one is \emph{Human3.6M}~\cite{ionescu2013human3}, which contains video sequences of different human actors captured from 4 synchronized cameras. Following \cite{peng2021animatable}, we conduct experiments on 7 subject: S1, S5, S6, S7, S8, S9, and S11. Specifically, we train 7 MPS-NeRF models where each model is trained with 6 subjects and tested on the remaining 1 subject in a leave-one-out cross-validation setup. 
{\annotated As in AniNeRF~\cite{peng2021animatable}, we use three views (\#0, \#1, \#2) out of the four as the input to our method and all four views for supervision during training. For testing of novel view synthesis, we input 3 views (\#0, \#1, \#2) and test on view \#3 for each frame. For testing of novel pose synthesis, we choose 3 views (\#0, \#1, \#2) of one pose as our input, construct the canonical human NeRF, and deform it to all target novel poses at view \#3 (\emph{i.e.,} novel pose synthesis at novel view). The detailed frame number used in the test set is presented in  Table~\ref{tab:frames_h36m}. We train and test our method using fitted SMPL parameters and image masks provided by \cite{peng2021animatable} which are obtained using \cite{joo2018total} and \cite{gong2018instance}, respectively.}

To evaluate our method on a larger people collection, 
we construct another dataset using textured 3D human meshes of 30 randomly-selected subjects from the \emph{THuman} dataset~\cite{zheng2019deephuman}. Each person has about 30 meshes with different poses, from which 20 are randomly chosen to construct our multi-view, multi-pose image dataset. We randomly split the 30 subjects into a training set with 25 subjects and a test set with the remaining 5 subjects. 
For each person and each pose, we render 24 images from different camera viewpoints. All the cameras point to body center with their azimuth and elevation angles evenly-sampled in $[0^\circ, 360^\circ]$ and $[0^\circ, 35^\circ]$, respectively. 
The reconstructed SMPL parameters provided by the THuman dataset~\cite{zheng2019deephuman} are used.

\begin{table}[t!]
	\centering
	\small
	\caption{Number of frames used to evaluate the novel-view and novel-pose synthesis tasks on Human3.6M.\label{tab:frames_h36m}}
	\begin{tabular}{c|ccccccc}
		\hline
		& S1 & S5  & S6 & S7 & S8  & S9 & S11 \\
		\hline
		Novel view & \!150\! & \!250\! & \!150\! & \!300\! & \!250\! & \!260\! & \!200\! \\
		Novel pose & 49 & 127 & 83 & 200 & 87 & 133 & 82 \\
		\hline
	\end{tabular}
\end{table}

\vspace{4pt}
\noindent\textbf{Evaluation metric} 
PSNR and SSIM metrics are used for  quantitative evaluation.  Instead of directly calculating PSNR and SSIM for the whole image, we follow AniNeRF~\cite{peng2021animatable} and NeuralBody~\cite{peng2021neural} to project the 3D bounding box of a body onto image plane to obtain a 2D mask and only calculate PSNR and SSIM in the masked region.

\begin{table*}[t!]
	\centering
	\small
	\caption{Comparison of our method with NB~\cite{peng2021neural}, AniNeRF~\cite{peng2021animatable} on the Human3.6M dataset. Best and second best results are marked with bold and underline, respectively. \label{tab:novel_view}}
	\setlength{\tabcolsep}{1.0 mm}{
				\begin{tabular}{c|cccccc|cccccc}
					\toprule
					\multicolumn{1}{c|}{\multirow{3}{*} {Subject}} & \multicolumn{6}{c|}{Novel View Synthesis} & \multicolumn{6}{c}{Novel Pose Synthesis}\\
					& \multicolumn{3}{c}{PSNR} & \multicolumn{3}{c|}{SSIM} & \multicolumn{3}{c}{PSNR} & \multicolumn{3}{c}{SSIM} \\
					\cline{2-13}
					& ~~~NB~~~ & \!\!AniNeRF\!\!  & ~Ours~ & ~~~NB~~~ & \!\!AniNeRF\!\!  & ~Ours~ & ~~~NB~~~ & \!\!AniNeRF\!\!  & ~Ours~ & ~~~NB~~~ & \!\!AniNeRF\!\!  & ~Ours~ \\
					\hline
					S1    & \underline{22.87}  & 22.05 & \textbf{25.40} & \underline{0.897} & 0.888 & \textbf{0.926} &\textbf{22.11}	&21.37	& \underline{21.87} &\underline{0.879}	&0.868	& \textbf{0.880} \\
					S5    & \textbf{24.6}  & 23.27 & \underline{24.30}  & \textbf{0.917} & 0.892 & \underline{0.908} & \textbf{23.51} & \underline{22.29} & 21.49 & \textbf{0.897} & \underline{0.875} & 0.871 \\
					S6    & \underline{22.82} & 21.13 & \textbf{23.94} & \underline{0.888} & 0.854 & \textbf{0.893} & \underline{23.52} & 22.59 & \textbf{23.63} & \underline{0.889} & 0.884 & \textbf{0.891} \\
					S7    & \underline{23.17} & 22.50 & \textbf{24.27} & \textbf{0.914} & 0.890 & \underline{0.911} & \textbf{22.33} & \underline{22.22} & 21.88 & \textbf{0.889} & \underline{0.878} & 0.868 \\
					S8    & 21.72 & \underline{22.75} & \textbf{23.66} & 0.894 & \underline{0.898} & \textbf{0.920} & 20.94 & \textbf{21.78} & \underline{21.15} & 0.876 & \underline{0.882} & \textbf{0.888} \\
					S9    & 24.28 & \textbf{24.72} & \underline{24.55} & \textbf{0.910} & \underline{0.908} & 0.899 & 23.04 & \textbf{23.72} & \underline{23.33} & \underline{0.884} & \textbf{0.886} & 0.875 \\
					S11   & 23.70 & \underline{24.55} & \textbf{25.12} & 0.896 & \underline{0.902} & \textbf{0.913} & \underline{23.72} & \textbf{23.91} & 23.53 & 0.884 & \underline{0.889} & \textbf{0.891} \\
					Average & \underline{23.31} & 23.00    & \textbf{24.46} & \underline{0.903} & 0.890  & \textbf{0.910} & \textbf{22.74} & \underline{22.55} & 22.41 & \textbf{0.885} & 0.880  & \underline{0.881} \\
					
					\bottomrule
				\end{tabular}%
		}
	\end{table*}%
\begin{figure*}[th!]
	\centering
	\includegraphics[width=0.99\linewidth]{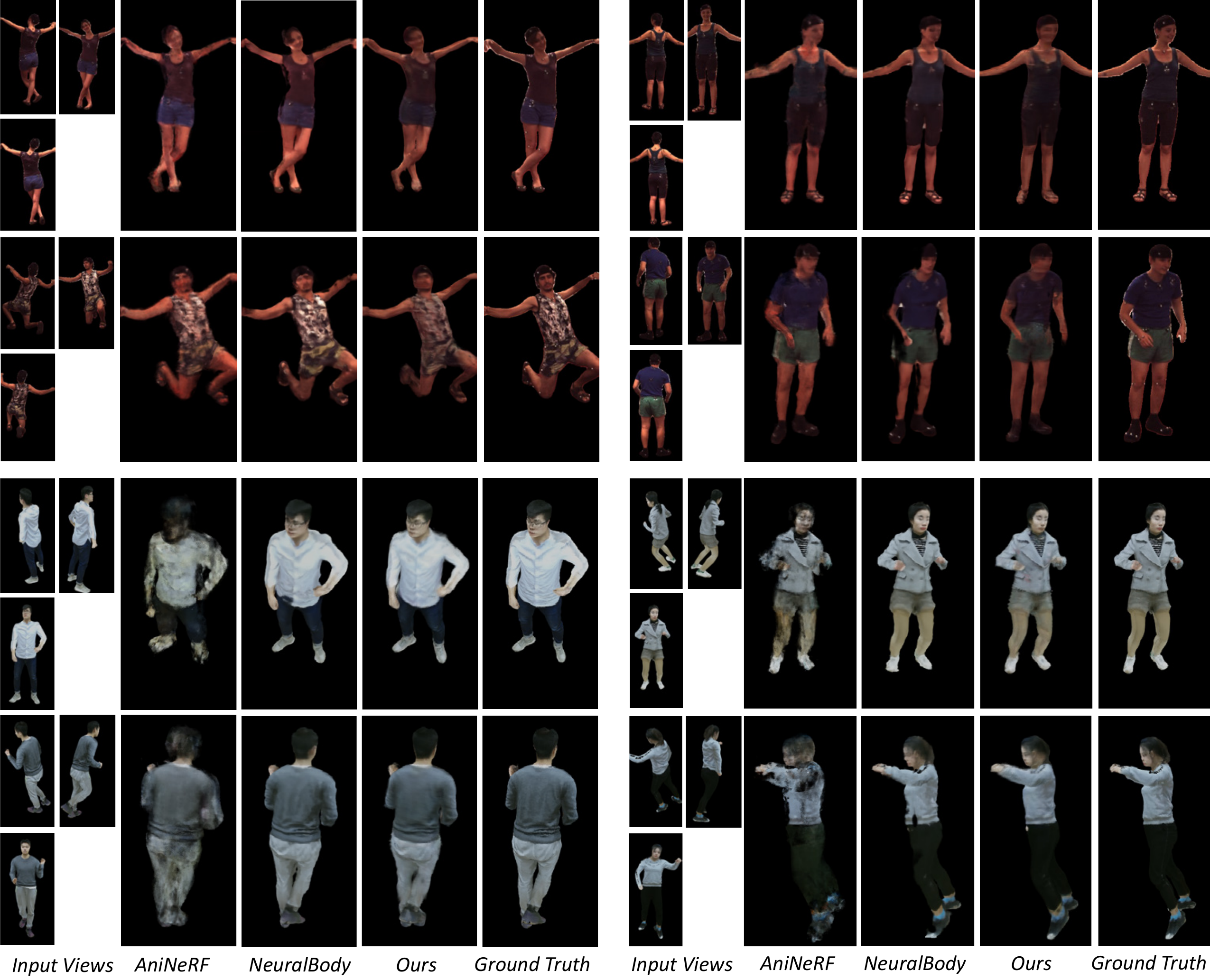}
	\captionsetup{type=figure,font=small}
	\caption{
		Novel view synthesis results on Human3.6M (top two rows) and THuman (bottom two rows). Note that the three images on the left are only used by our method as input to render these \emph{unseen} subjects at inference time. NeuralBody~\cite{peng2021neural} and AniNeRF~\cite{peng2021animatable} are person-specific models  which only need camera parameters to render a novel view of these \emph{trained} subjects.
	}
	\label{fig:novel_view}
\end{figure*}

\begin{figure*}[th!]
	\centering
	\includegraphics[width=1\linewidth]{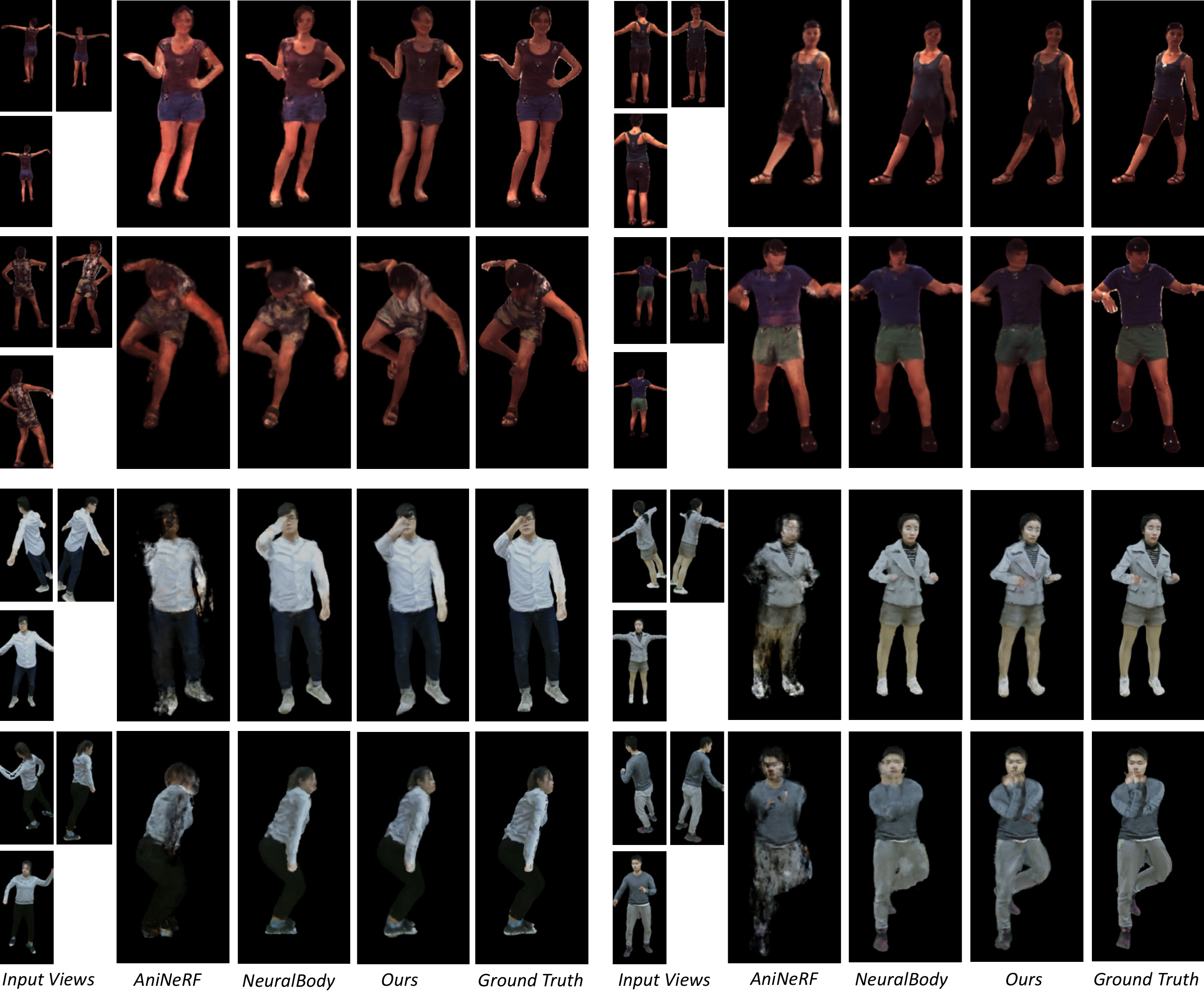}
	\vspace{-12pt}
	\captionsetup{type=figure,font=small}
	\caption{
		Novel pose synthesis results on the Human3.6M (top two rows) and THuman (bottom two rows) datasets. Note that the three images on the left are only used by our method as input to render these \emph{unseen} subjects at inference time. NeuralBody~\cite{peng2021neural} and AniNeRF~\cite{peng2021animatable} are person-specific models which only need camera and pose parameters to render these \emph{trained} subjects.\label{fig:novel_pose}
	}
\end{figure*}

\subsection{Results and Comparison with Prior Art}

\noindent\textbf{Competing methods.}
To our knowledge, MPS-NeRF is the first person-agnostic framework for novel-view and novel-pose human synthesis using multiview images.
Hence, for reference purpose, we compare our method with recent person-specific models NeuralBody (NB)~\cite{peng2021neural} and Animatable NeRF (AniNeRF)~\cite{peng2021animatable}.
For these methods, one model is trained and tested on a single subject, so 7 models on Human3.6M and 5 models on THuman are used in our experiments. For the evaluation on Human3.6M, we report the numerical and visual results of NB and AniNeRF provided by the authors. To evaluate the methods on THuman, we train the two methods using the released source codes.

Note that for fair comparison, \textbf{\emph{all experimental configurations such as input SMPL parameters, image masks, input views, test image sets, and evaluation protocols are made identical for all the methods}}.

\begin{table}[t!]
	\centering
	\small
	\caption{Comparison of NB~\cite{peng2021neural}, AniNeRF~\cite{peng2021animatable}, and our method on the THuman dataset.}
	\label{tab:thuman}%
	\setlength{\tabcolsep}{2.5mm}{
				\begin{tabular}{ccccc}
					\toprule
					\multicolumn{1}{c}{\multirow{2}{*} {Method}} & \multicolumn{2}{c}{Novel View} & \multicolumn{2}{c}{Novel Pose} \\
					\cmidrule{2-5}    & PSNR & SSIM  & PSNR & SSIM  \\
					\midrule
					NB    & 24.86 & 0.929 & 23.36 & 0.903 \\
					AniNeRF & 20.10  & 0.841 & 17.25 & 0.791  \\
					Ours  & \textbf{25.63} & \textbf{0.935} & \textbf{23.92} & \textbf{0.911} \\
					\bottomrule
				\end{tabular}%
	}
\end{table}

\vspace{4pt}
\noindent\textbf{Novel view synthesis results.}
In our experiments on Human3.6M, all the methods are evaluated on the testing splits of NB and AniNeRF. 
Table~\ref{tab:novel_view} presents the quantitative comparison of our MPS-NeRF with the other two methods.  Although all the testing persons are unseen to MPS-NeRF, it yields better novel view synthesis results than the other two approaches (\eg, $1$dB higher in average PSNR).
Figure~\ref{fig:novel_view} shows the visual result comparison of our method and the others. It can be observed that our method generalizes well to the novel testing persons in the view synthesis task. Compared to NB and AniNeRF, our results exhibit less texture distortions. 

To evaluate the methods on THuman, we use images with three evenly-spaced  azimuth angles 0$^\circ$, 120$^\circ$ and 240$^\circ$ as the input for all methods and test on another 8 views. With these 3 views, NB and AniNeRF are trained on all 20 poses of each of the 5 testing person. In contrast, we train a single MPS-NeRF model on 25 training persons each with 20 poses, and use all 24 views as supervision. 
Table~\ref{tab:thuman} shows the numerical results of different methods on the test set. Again, our method yields better results than both NB and AniNeRF in terms of the average PSNR and SSIM. AniNeRF clearly underperforms in this case. As the pose number for each person is limited and there is no global body rotation, AniNeRF struggles to predict a reasonable novel view image.

Figure~\ref{fig:novel_view} presents the qualitative results of different methods. Visually inspected, NB and our method are able to generate reasonable results with comparable quality.

\begin{figure}[t!]
	\centering
	\includegraphics[width=1.0\columnwidth]{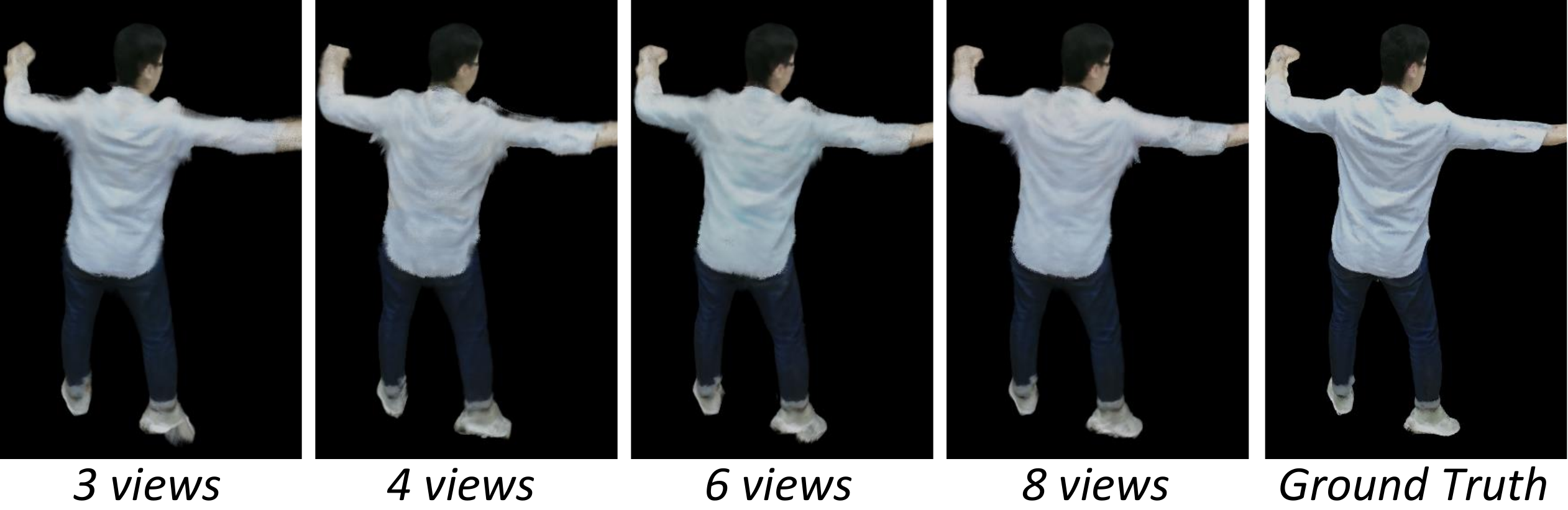}
	\vspace{-13pt}
	\captionsetup{type=figure,font=small}
	\caption{
		Visual results with different input view numbers.\label{fig:ablation_view_num}
	}
\end{figure}

\begin{figure}[t!]
	\centering
	\includegraphics[width=1.0\columnwidth]{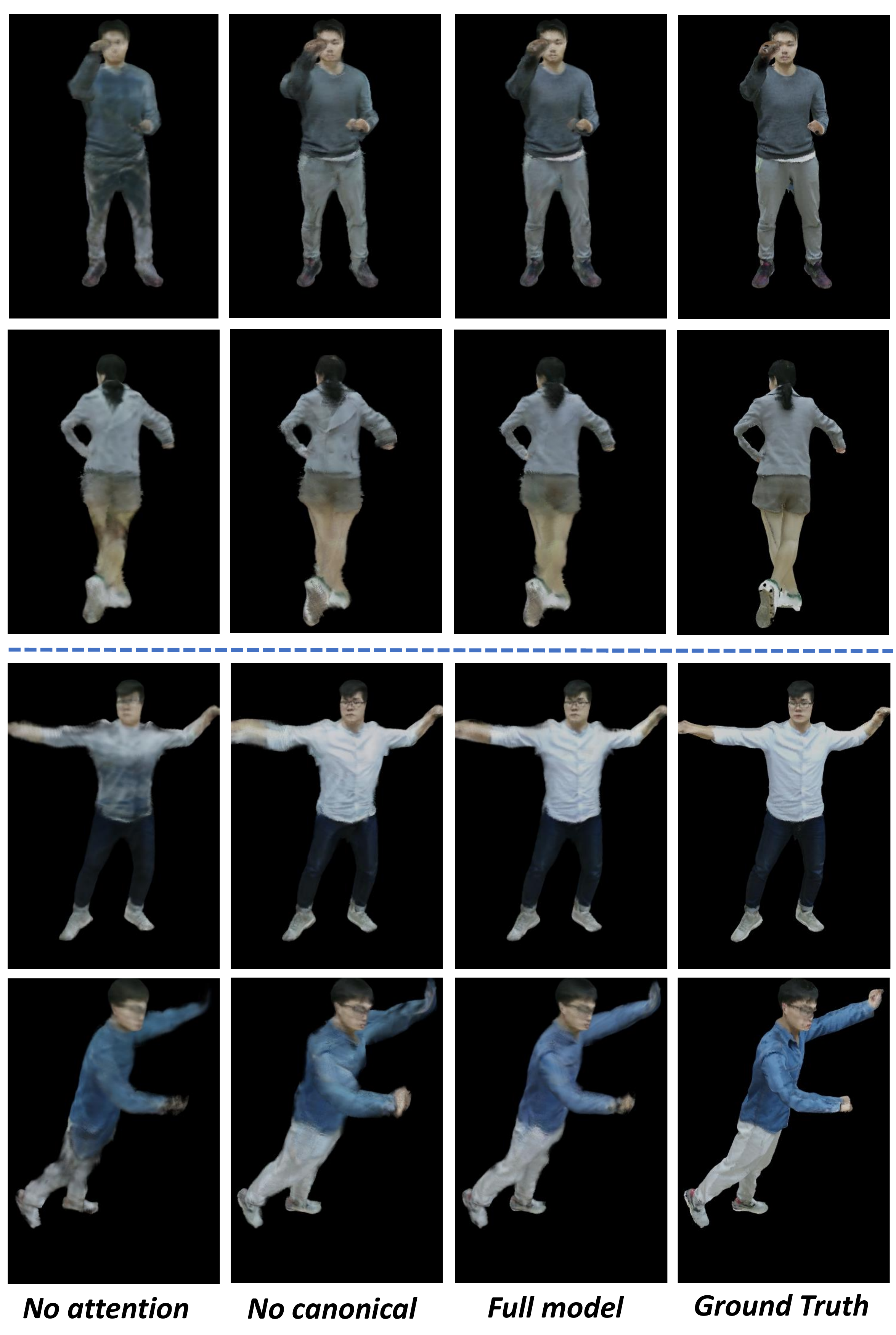}
	\vspace{-15pt}
	\captionsetup{type=figure,font=small}
	\caption{Impact of canonical space and attention-based feature fusion on novel view (top two rows) and novel pose (bottom two rows) synthesis tasks.\label{fig:ablation_can_trans_novel_view}
	}
\end{figure}

\begin{figure}[t!]
	\centering
	\includegraphics[width=1.0\columnwidth]{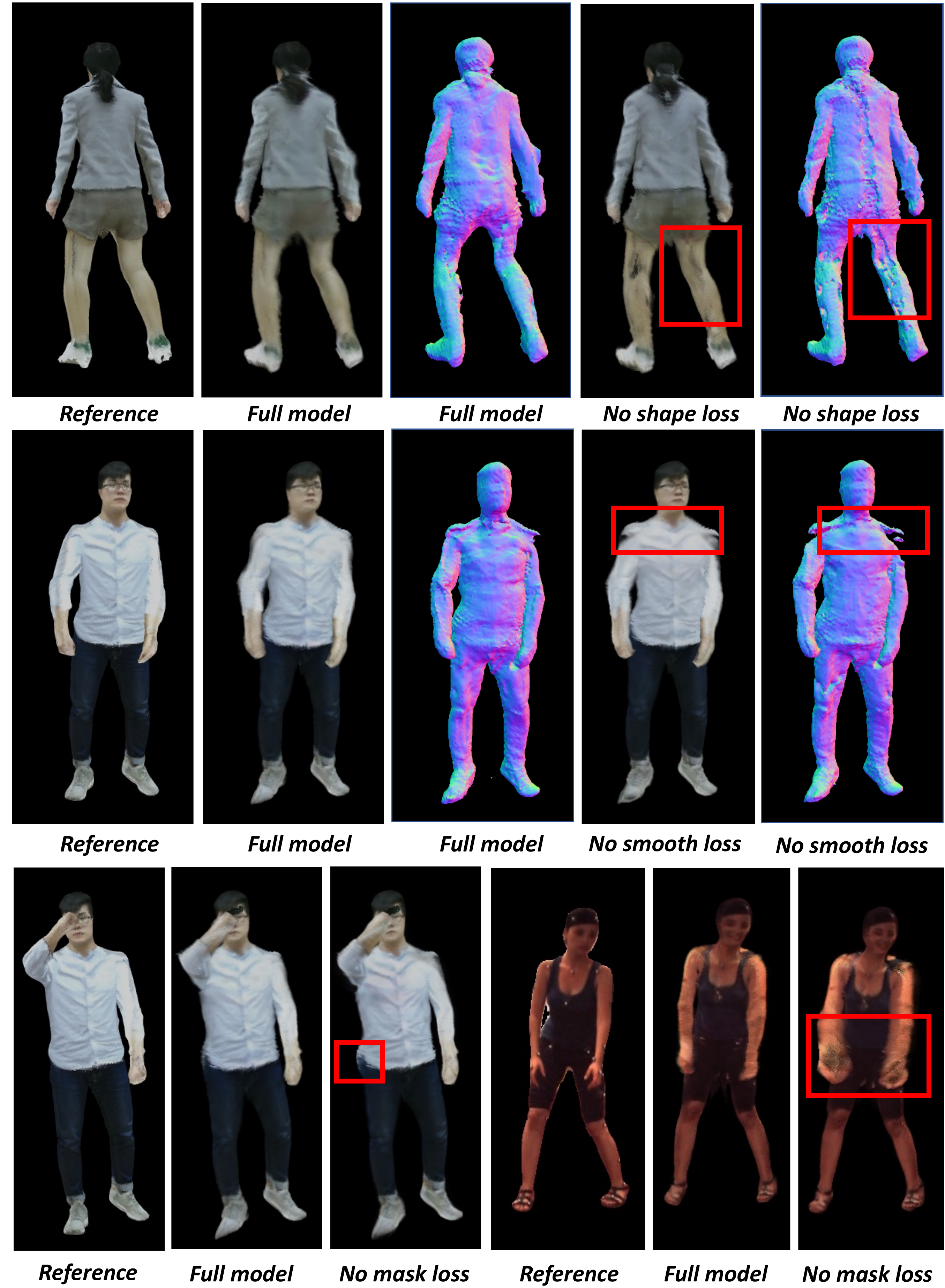}
	\vspace{-15pt}
	\captionsetup{type=figure,font=small}
	\caption{{\annotated Visual examples of loss term ablation study.}\label{fig:loss_ablation}
	}
\end{figure}

\vspace{4pt}
\noindent\textbf{Novel pose synthesis results.} For the novel pose synthesis task on Human3.6M, we evaluate the results of different methods with the testing view and poses from the test sets of NB and AniNeRF. For our MPS-NeRF, we choose one pose from the training set of NB and AniNeRF to synthesize the novel pose of an unseen testing person. The quantitative results are presented in Table~\ref{tab:novel_view}. It shows that all three methods performed comparably, with our result {marginally worse} than NB and AniNeRF (\eg, average PNSR 22.41dB from our method \emph{vs.} 22.74dB and 22.55dB from NB and AniNeRF, respectively). This indicates that our trained model generalizes well to not only unseen person but also various poses. Figure~\ref{fig:novel_pose} shows some visual results of different methods. While it appears that our results are slightly more blurry than NB and AniNeRF, it generally contains less high-frequency artifacts.

In the novel pose synthesis experiments on THuman, we also use three input views as in the novel view synthesis task. NB and AniNeRF are trained on all the 20 poses of each person. For evaluation, we take another 5 poses from the THuman dataset and render a novel pose test set for each subject. Our method is trained on 25 training subjects each with 20 poses, and tested on the same test set as NB and AniNeRF. The quantitative results presented in Table~\ref{tab:thuman} shows that our method performed best. It is slightly better than NB (average PSNR 23.92dB \emph{vs.} 23.36dB). Again, AniNeRF performed poorly due to limited training poses. Some visual results are presented in Figure.~\ref{fig:novel_pose}.

\begin{figure*}[t!]
\centering
\includegraphics[width=1.0\linewidth]{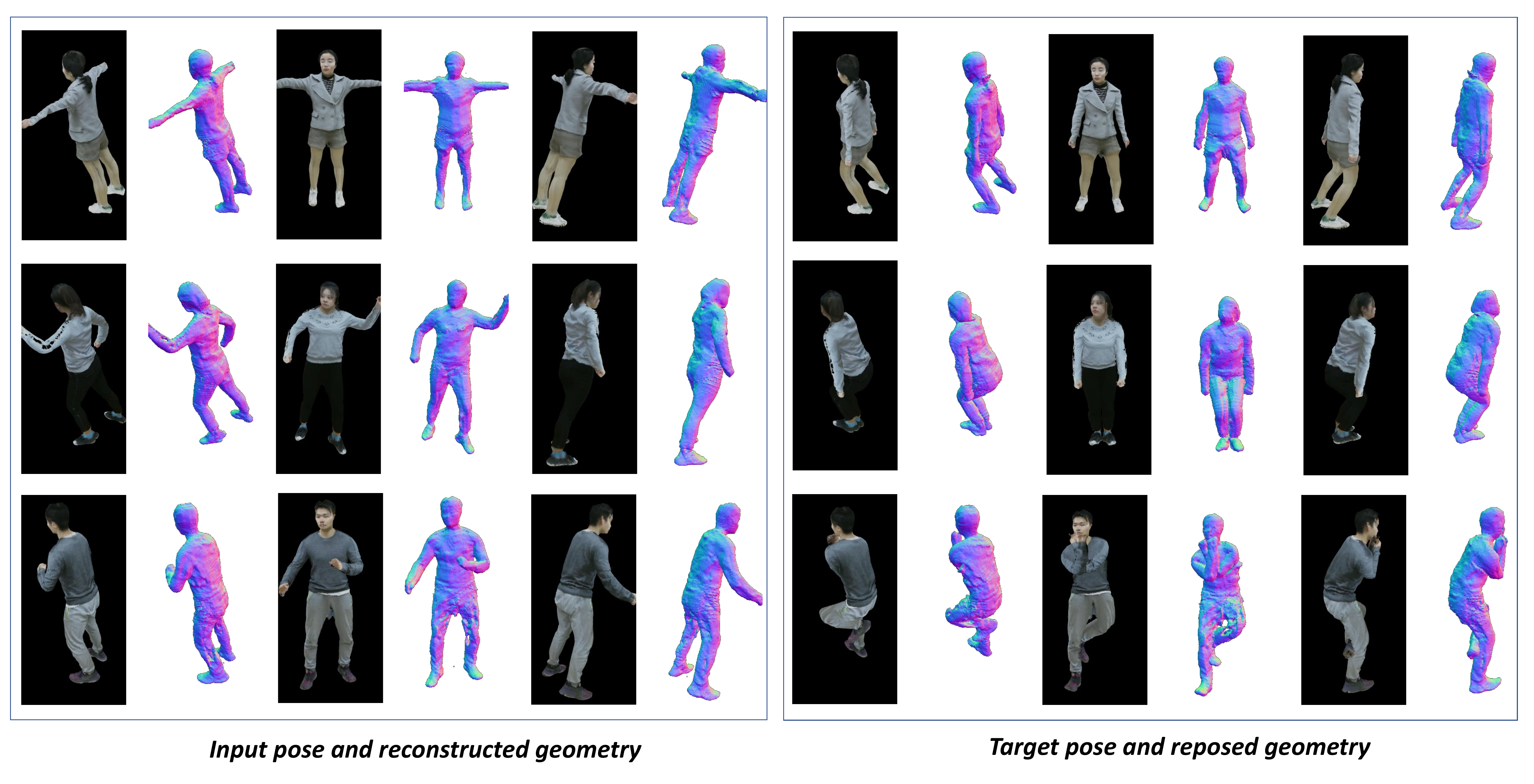}
\vspace{-18pt}
\captionsetup{type=figure,font=small}
\caption{Proxy 3D shapes reconstructed by our method. The meshes are extracted by running the Marching Cubes algorithm on the binarized volume density, and we did not apply any post-processing such as mesh smoothing.\label{fig:geometry}}
\end{figure*}

\subsection{Analysis and Ablation study}
We further test MPS-NeRF with more input views on the THuman dataset and analyze their impact. We also conduct ablation studies to validate the efficacy of different modules.

\vspace{4pt}
\noindent\textbf{Impact of input view number.}
In this experiment, we train and test our method using different numbers of input views from 3 to 12.
Table~\ref{tab:ablation_view_num} shows that the performance of our method gradually increases with more input views and gets saturated when the number is greater than 8. One visual example is presented in Fig.~\ref{fig:ablation_view_num}, which shows that the noise is gradually suppressed and boundary becomes sharper and more accurate with an increasing input view number.

\begin{table}[t!]
	\centering
	\small
	\caption{Quantitative results with different input view numbers.\label{tab:ablation_view_num}}	
	\setlength{\tabcolsep}{1.5mm}{
				\begin{tabular}{ccccc}
					\toprule	\multicolumn{1}{c}{\multirow{2}{*} {View Number}} & \multicolumn{2}{c}{Novel View} & \multicolumn{2}{c}{Novel Pose} \\
					\cmidrule{2-5}    & PSNR & SSIM  & PSNR & SSIM  \\
					\midrule
					3     & 25.63 & 0.935 & 23.92 & 0.911 \\
					4     & 25.84 & 0.938 & 23.95 & 0.912 \\
					6     & 26.46 & 0.943 & 24.25 & 0.916 \\
					8     & 27.37 & 0.951 & 24.45 & 0.92 \\
					12    & \textbf{27.77} & \textbf{0.953} & \textbf{24.56} & \textbf{0.921} \\
					\bottomrule
				\end{tabular}%
	}
\end{table}

\begin{table}[t!]
	\centering
	\small
	\caption{Ablation study on {\annotated our canonical space design, feature attention strategy, and loss terms}.\label{tab:ablation_can_trans}}
	\setlength{\tabcolsep}{1.5mm}{
				\begin{tabular}{lcccc}
					\toprule	\multicolumn{1}{c}{\multirow{2}{*} {}} & \multicolumn{2}{c}{Novel View} & \multicolumn{2}{c}{\tabincell{c}{Novel Pose}} \\
					\cmidrule{2-5}    & PSNR & SSIM  & PSNR & SSIM  \\
					\midrule
					w/o canonical & 23.34 & 0.913 & 21.77 & 0.888 \\
					w/o attention & 22.05 & 0.907 & 21.18 & 0.888 \\
					{\annotated w/ single attention} & {\annotated 25.29} & {\annotated 0.932}  & {\annotated 23.63} & {\annotated 0.909} \\
					\midrule
					{\annotated w/o shape loss} & {\annotated 25.16} & {\annotated 0.928}  & {\annotated 23.57} & {\annotated 0.905} \\
					{\annotated w/o smoothness loss} & {\annotated 25.38} & {\annotated 0.933}  & {\annotated 23.64} & {\annotated 0.909} \\
					\midrule
					Our full model & \textbf{25.63} & \textbf{0.935} & \textbf{23.92} & \textbf{0.911} \\
					\bottomrule
				\end{tabular}%
		}
	\end{table}

\vspace{4pt}
\noindent\textbf{Impact of canonical space.}
To validate the importance of using a canonical space, we remove it in our implementation and retrain our model. Note that in this setup, our method is very similar to the image-conditioned NeRF methods for generic scenes~\cite{yu2021pixelnerf,trevithick2021grf}. As shown in Table~\ref{tab:ablation_can_trans}, the performance drops significantly for both the novel-view and novel-pose synthesis tasks, which indicates that our canonical space design is critical for generalizable human rendering. \textcolor{black}{We further show the visual comparison in Fig.~\ref{fig:ablation_can_trans_novel_view}. As can be observed, the results without a canonical space contain more unwanted artifacts. Moreover, the predicted human boundary is clearly erroneous for some cases (\eg, see the arm in the third row).}

\vspace{4pt}
\noindent\textbf{Impact of attention-based feature fusion.} We also verify the efficacy of our attention-based feature fusion scheme by replacing it with a simple feature average pooling strategy. The performance also drops dramatically as shown in Table~\ref{tab:ablation_can_trans}, demonstrating that multiview feature fusion cannot be handle naively and our feature fusion method is substantial for good performance.

Figure \ref{fig:ablation_can_trans_novel_view} shows that without our attention-based fusion, the synthesized images are blurry and suffer from severe color distortions.
{\annotated In our feature fusion scheme, we designed two attention MLPs to generate two features for geometry and color, respectively. Here, we compare with a single attention block which generates one fused feature, and Table~\ref{tab:ablation_can_trans} shows the superior performance of our design.
}

{\annotated
\vspace{4pt}
\noindent\textbf{Impact of different loss terms.}
The ablation study on our loss terms are shown in Table~\ref{tab:ablation_can_trans}, with some typical visual results presented in Fig.~\ref{fig:loss_ablation}. The shape loss imposes strong regularization on geometry, without which obvious geometry distortions may occur. 
The smoothness loss also improves the geometry by diminishing noisy and disconnected regions. The improved geometry leads to better rendering quality.
For the mask loss, we found it brings no noticeable benefit in terms of average PSNR and SSIM metrics. However, dropping it leads to some large shape distortions in some rare cases, as shown in Fig.~\ref{fig:loss_ablation}.
}

\begin{figure}[!t]
	\centering
	\includegraphics[width=1.0\columnwidth]{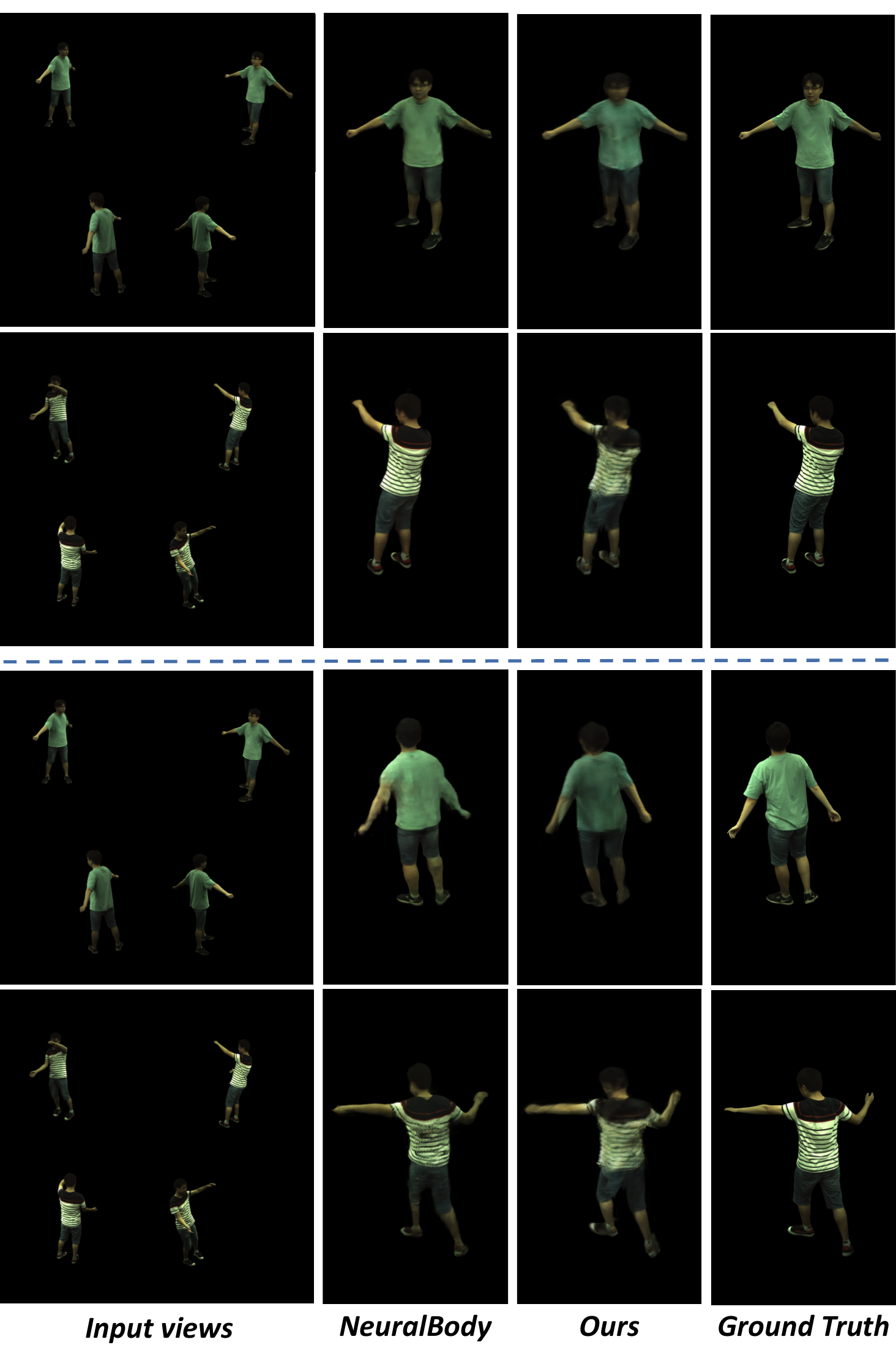}
	\vspace{-15pt}
	\captionsetup{type=figure,font=small}
	\caption{Results of our method tested on two subjects from the ZJU-MoCap dataset~\cite{peng2021neural}. In this dataset, the actors make constant body spinning during video shooting, which greatly eases the novel-view synthesis learning task for \emph{person-specific} models. Our method still generates reasonable results for both novel-view and novel-pose synthesis despite the person is \emph{unseen} during training.\label{fig:zju}
	}
\end{figure}

\subsection{Reconstructed 3D Shapes}
	
Although the NeRF-based representation does not explicitly recover 3D geometry, we can still extract proxy 3D shapes using the density field predicted by the network. Specifically, we first define a 3D human bounding box based on the target pose SMPL model which has a size of $2m\times 2m\times 2m$, and discretize it to a $256 \times 256 \times 256$ voxel grid.
Then we deform the voxel centers from the target space to the canonical space, and evaluate their volume densities using our canonical NeRF network. Finally, the density values are binarized and the Marching Cubes algorithm~\cite{lorensen1987marching} is applied to extract a mesh. 
Figure~\ref{fig:geometry} presents the extracted proxy 3D shapes for the input images as well as the reposed  target shapes using our deformation scheme. Note that we did not apply any post-processing on the extracted 3D shapes such as the Gaussian smoothing used in AniNeRF~\cite{peng2021animatable}.

\subsection{More results on ZJU-MoCap Dataset}

We further tested our method on the ZJU-MoCap dataset~\cite{peng2021neural}, where we train on 6 subjects using 4 views as input. Figure~\ref{fig:zju} shows the results of our method on an unseen testing person. 
Our method still generates reasonable results for both the novel-view and novel-pose synthesis tasks, despite the person is not in the training set. 
For a reference, we also present the visual results of NeuralBody~\cite{peng2021neural}. Note that in this dataset, the actors make constant body spinning during video shooting, unlike the Human3.6M dataset where the body movements are more natural. This greatly eases the novel-view synthesis learning task for person-specific models. Therefore, NeuralBody can generate remarkable novel view synthesis results. Still, our novel pose synthesis results are comparable to NeuralBody, despite the person is unseen to our model.

\section{Conclusion}
In this paper, we proposed a novel-view and novel-pose human synthesis approach which can be generalizable for unseen persons with sparse multiview images as input. Our key idea is to leverage a canonical-space NeRF and a volume deformation scheme derived by human body parametric model to achieve better generalizability.
Our method is simple but works  well as demonstrated by the extensive experiments on both real and synthetic datasets. We hope that our method can serve as a strong baseline model for generic 3D human rendering.

Our current deformation scheme cannot handle very loose clothing and garments with complex geometry (\eg, a long coat or opened jacket). Our future work will be devoted to designing a generalizable, learning-based deformation scheme to better handle such cases.
{\annotated The current rendering quality is far from perfect, and the synthesized images are still blurry (despite this is also partially due to the low-quality input). In this work, we simply use a naive $L_2$ rendering loss for training following NB and AniNeRF. Other sharpness-promoting losses can be applied in future.}


\bibliographystyle{IEEEtran}
\bibliography{references}

\ifpeerreview \else

\begin{IEEEbiography}[{\includegraphics[width=1.0in,clip,keepaspectratio]{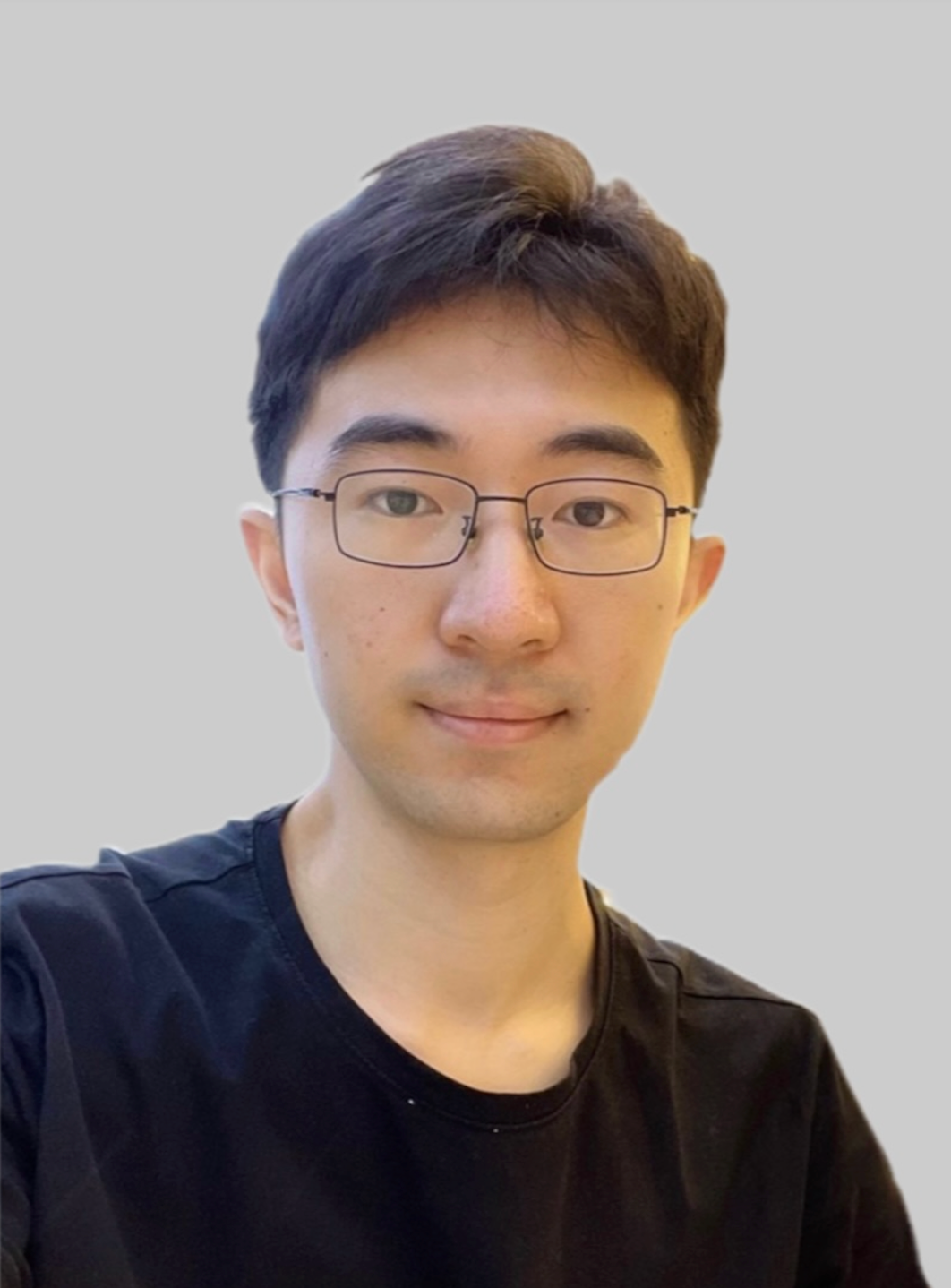}}]{Xiangjun Gao}
is currently a master student of Computer Science at Beijing Institute of Technology, advised by Assitant Prof. Yuwei Wu and Prof. Yunde Jia. He received B.E. degree in Computer Science and Engineering from Beijing Institute of Technology in 2020. His research interests include differential rendering and virtual avatar .
\end{IEEEbiography}

\begin{IEEEbiography}[{\includegraphics[width=1.0in,clip,keepaspectratio]{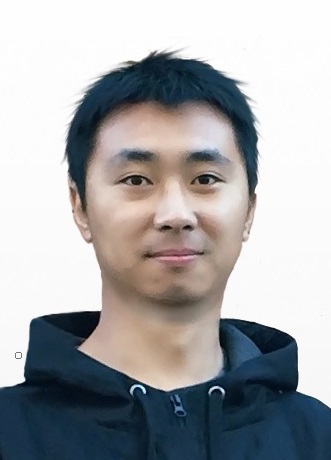}}]{Jiaolong Yang} is currently a senior researcher at Microsoft Research Asia, Beijing, China. He received the dual Ph.D. degrees in Computer Science and Engineering from the Australian National University and Beijing Institute of Technology in 2016. His research interests include 3D vision for human face and body. He received the outstanding PhD thesis award from China Society of Image and Graphics in 2017 and the best paper award at IEEE VR/TVCG 2022.
\end{IEEEbiography}

\begin{IEEEbiography}[{\includegraphics[width=1.0in,clip,keepaspectratio]{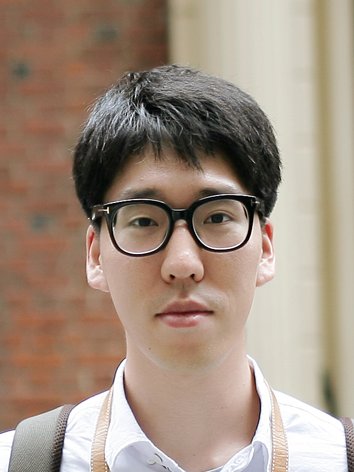}}]{Jongyoo Kim} is a senior researcher at Microsoft Research Asia. received his B.S. degree, M.S. degree and Ph.D. degree in Electrical and Electronic Engineering from Yonsei University, Seoul, Korea in 2011, 2013 and 2018, respectively. His research interests include 2D/3D computer vision, computer graphics, and perceptual image/video processing. He was a recipient of the Global PhD Fellowship by National Research Foundation of Korea from 2011 to 2016.
\end{IEEEbiography}

\begin{IEEEbiography}[{\includegraphics[width=1.0in,clip,keepaspectratio]{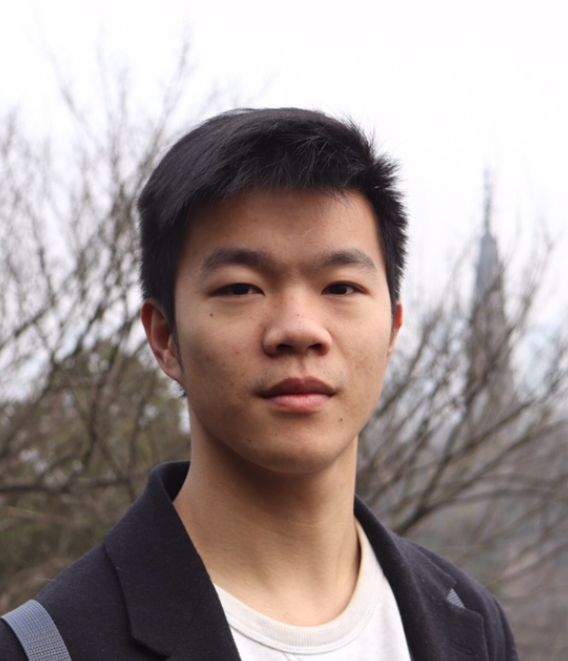}}]{Sida Peng} is currently a Ph.D. student of computer science at Zhejiang University, advised by Dr. Xiaowei Zhou. He received B.E. degree in information engineering from Zhejiang University in 2018. His research interests include 3D reconstruction and object pose estimation.
\end{IEEEbiography}

\begin{IEEEbiography}[{\includegraphics[width=1.0in,clip,keepaspectratio]{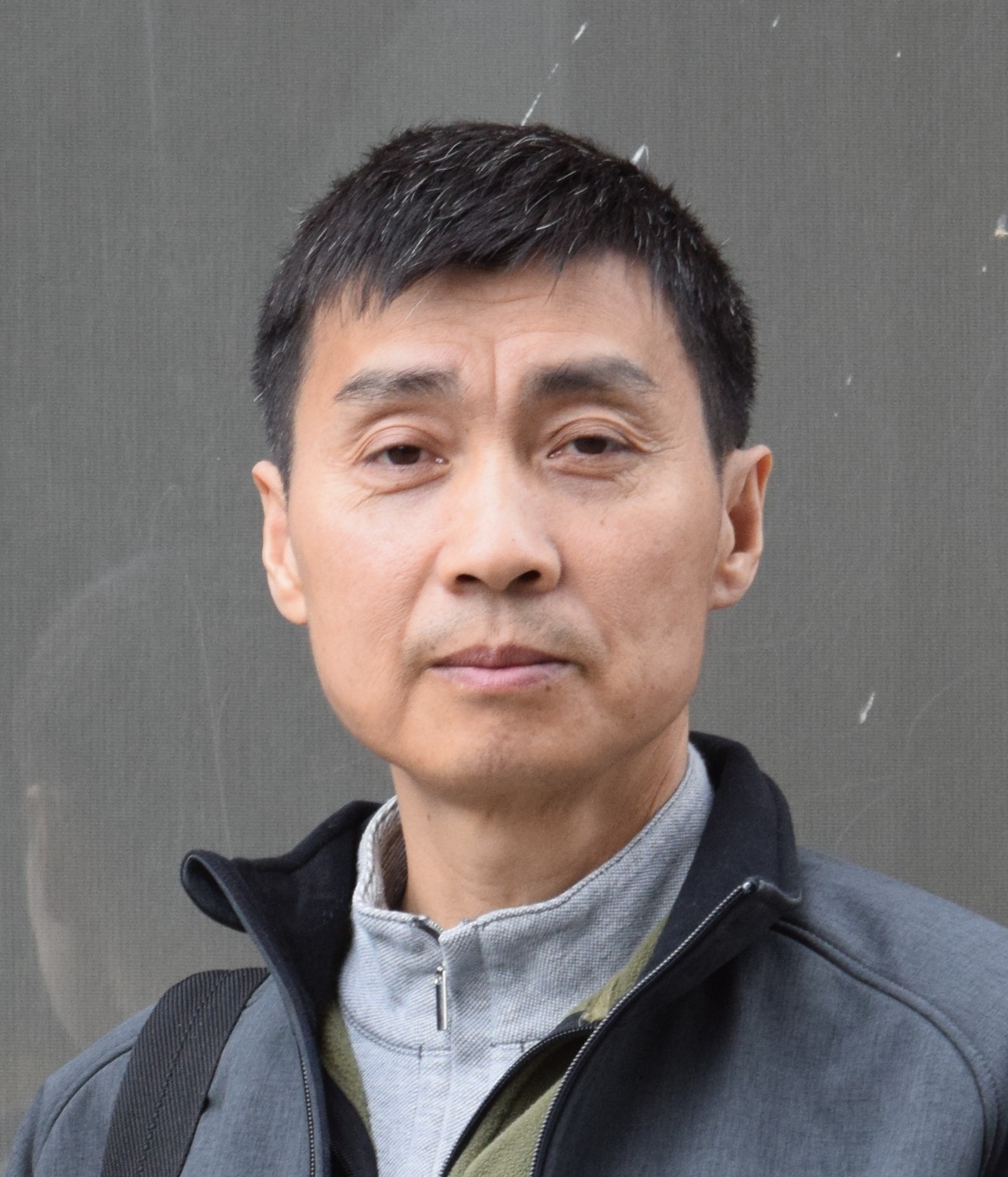}}]{Zicheng Liu} is currently a partner research manager at Microsoft.  His research interests include computer vision, vision-language learning, and machine learning. He received a Ph.D. in Computer Science from Princeton University, a M.S. in Operational Research from the Institute of Applied Mathematics, Chinese Academy of Science, and a B.S. in Mathematics from Huazhong Normal University, China. 
He served as the chair of IEEE CAS Multimedia Systems and Applications technical committee from 2015-2017. He was a distinguished lecturer of IEEE CAS from 2015-2016. He served as a member of the Steering Committee of IEEE Transactions on Multimedia. He was a technical co-chair of 2010 and 2014 IEEE International Conference on Multimedia and Expo (ICME), and a general co-chair of 2012 and 2022 IEEE Visual Communication and Image Processing (VCIP). He is the Editor-in-Chief of the Journal of Visual Communication and Image Representation. He is a fellow of IEEE.
\end{IEEEbiography}

\begin{IEEEbiography}[{\includegraphics[width=1.0in,clip,keepaspectratio]{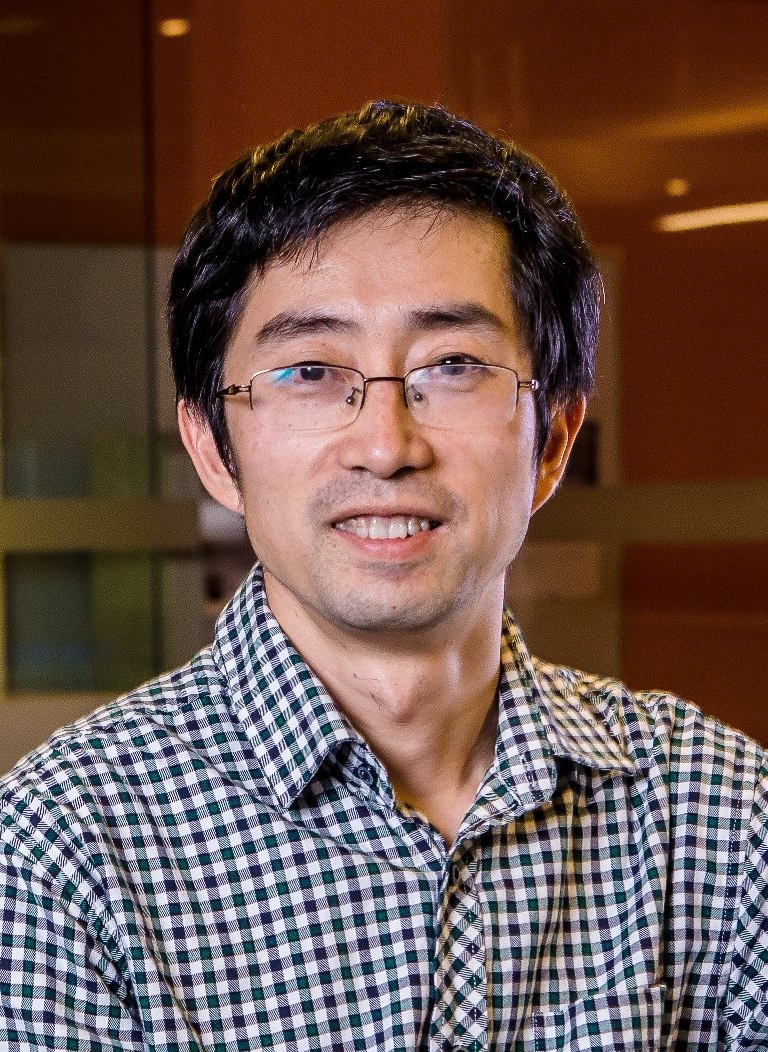}}]{Xin Tong}
is currently a partner research manager at internet graphics group of Microsoft Research Asia. His research interests include computer graphics and computer vision. Xin received his Ph.D. in computer science from Tsinghua University. He served as associate editor of IEEE TVCG, ACM TOG, and CGF and was regularly invited to be committee member of international graphics conferences such as ACM SIGGRAPH and SIGGRAPH ASIA, Eurographics, and so on. He is the associate editor of IEEE CG\&A.
\end{IEEEbiography}

\fi

\end{document}